\documentclass{article}

\usepackage{arxiv}

\usepackage[utf8]{inputenc} % allow utf-8 input
\usepackage[T1]{fontenc}    % use 8-bit T1 fonts
\usepackage{hyperref}       % hyperlinks
\usepackage{url}            % simple URL typesetting
\usepackage{booktabs}       % professional-quality tables
\usepackage{amsfonts}       % blackboard math symbols
\usepackage{nicefrac}       % compact symbols for 1/2, etc.
\usepackage{microtype}      % microtypography
\usepackage{graphicx}
\usepackage[numbers]{natbib}
\usepackage{doi}
\usepackage{amsmath}

\title{Named Entity Normalization Model Using\\ Edge Weight Updating Neural Network:\\ Assimilation Between Knowledge-Driven Graph\\ and Data-Driven Graph}

\date{} 					% Or removing it

\author{ Sung Hwan Jeon\\
	Department of Industrial Engineering\\
	Seoul National University\\
	1, Gwanak-ro, Gwanak-gu, Seoul, Republic of Korea \\
	\texttt{sjeon@dm.snu.ac.kr} \\
	\And
	Sungzoon Cho \\
	Department of Industrial Engineering and Institute for Industrial Systems Innovation\\
	Seoul National University\\
	1, Gwanak-ro, Gwanak-gu, Seoul, Republic of Korea \\
	\texttt{zoon@snu.ac.kr} \\
}

\renewcommand{\shorttitle}{A Preprint for Neural Processing Letters}

\hypersetup{
	pdftitle={Named Entity Normalization Model Using Edge Weight Updating Neural Network: Assimilation Between Knowledge-Driven Graph and Data-Driven Graph},
	pdfsubject={cs.AI},
	pdfauthor={Sung Hwan Jeon, Sungzoon Cho},
	pdfkeywords={Named Entity Normalization, Edge Weight Updating Neural Network, Text Mining in Bioinformatics, Text Mining in Finance, Named Entity Graph},
}

\begin{document}
\maketitle

\begin{abstract}
	Discriminating the matched named entity pairs or identifying the entities' canonical forms are critical in text mining tasks.
	More precise named entity normalization in text mining will benefit other subsequent text analytic applications.
	We built the named entity normalization model with a novel Edge Weight Updating Neural Network.
	Our proposed model when tested on four different datasets achieved state-of-the-art results.
	We, next, verify our model's performance on NCBI Disease, BC5CDR Disease, and BC5CDR Chemical databases, which are widely used named entity normalization datasets in the bioinformatics field.
	We also tested our model with our own financial named entity normalization dataset to validate the efficacy for more general applications.
	Using the constructed dataset, we differentiate named entity pairs.
	Our model achieved the highest named entity normalization performances in terms of various evaluation metrics.
\end{abstract}

% keywords can be removed

\section{Introduction}
\label{introduction}
The text mining technology is undergoing a rapid evolution thanks to the exponential growth in the number of text-rich documents available online, and as a result, it is being widely applied in a range of domains such as finance and bioinformatics.
Text mining aims to extract the information from documents to derive valuable insights.
Documents subject to analysis contain many named entities, which are proper names that denote unique objects such as organizations, products, persons, and locations.
The technique used to extract named entities from documents is called named entity recognition (NER, henceforth).
Furthermore, named entity normalization (NEN, henceforth) involves matching extracted named entities with homogeneous identity and is pivotal for text mining tasks.

More specifically, in the biomedical domain, disease names and chemicals in drugs often have different surface forms while sharing the same concept.
Types of named entities with different surface forms that share same concept can be divided into following categories: (1) synonyms, (2) abbreviations, (3) acronyms, (4) different combinations of punctuations and alphabets, (5) descriptive phrases, and (6) possible NER parsing errors.
For example, ``hepatomegaly" and ``liver enlarged" do not have matching strings but the two disease names have identical meanings, and thus, these two named entities are synonyms.
Biomedical named entities have a wide variety of different surface forms compared with entities from other text sources.
More accurate named entity normalization techniques will potentially improve the quality of downstream tasks.
Moreover, matching entity pairs such as ``International Business Machines" and ``IBM", which are examples of acronyms, are very critical in financial text mining applications.
Linking entities with the same identity enables accurate sentiment analysis on firms and products.
Furthermore, evaluation of news impacts on the stock market requires the connection between news articles and related firms.
Given the wide range of named entities in bioinformatics and finance documents, the total number of tokens to be calculated for text clustering and classification is enormous.

The early NEN models explored knowledge-based approaches.
Generating the rules for named entity matching based on domain knowledge is valid only for the dataset in which the corresponding rules are already created.
The rule-based models are not robust for the neologisms.
In order to overcome the disadvantage that the rule-based model is not robust, models based on machine learning have been introduced.
However, machine learning models are limited to specific fields such as bioinformatics NEN and chemical engineering NEN due to lack of NEN datasets in other domains.
Our research aims to construct fully automated NEN model that can be applied to various other domains.
To test our model's robustness on different domain, we also apply the NEN dataset in finance.

An automated named entity normalization model reduce the burden of hand-mined information extraction tasks.
Clear linkage between entities with different forms, such as abbreviations and acronyms, aid in more accurate sentiment analysis.
The named entity normalization model also benefits the creation of more comprehensible classifying and clustering documents.
The primary contributions of our study are (1) constructing better performing NEN model using an Edge Weight Updating Neural Network and (2) applying our proposed model to bioinformatics NEN and financial NEN tasks.

The proposed method, that is, the edge weight updating neural network, consists of four parts: (1) ground truth entity graph construction, (2) similarity-based entity graph construction, (3) edge weight updating neural network training, and (4) edge weight updating neural network inferencing.
The main concept behind the Edge Weight Updating Neural Network is to minimize the Ground Truth Entity Graph's edge weight distributions and the Similarity-Based Entity Graph's edge weight distributions.
By minimizing the edge weight distributions on the two graphs, entity embeddings capture more accurate information on semantic similarity between matching entities.

Our proposed model is evaluated on three widely used bioinformatics datasets (NCBI Disease, BC5CDR Disease, and BC5CDR Chemical) and its performance is compared with other cutting-edge models.
Furthermore, to validate the efficacy of our proposed model in general NEN tasks, we construct a financial NEN dataset with state-of-the-art NER using BERT \cite{devlin2018bert}.
Using the constructed dataset, we propose the deep learning model to solve more practical financial NEN tasks.
Out dataset incorporates major challenges in entity matching: (1) synonyms, (2) abbreviations, (3) acronyms, (4) different combinations of punctuations and alphabets, (5) descriptive phrases, and (6) possible NER parsing errors.
Compare with other recent NEN models, our proposed model shows higher accuracies in all datasets used in the experiments, and our model is tested with not only bioinformatics NEN datasets but also financial NEN datasets, which verifies the efficacy in general NEN tasks.

The remainder of this paper is organized as follows.
Section \ref{related_work} describes related work.
Section \ref{dataset} presents an overview of dataset we used for evaluations.
A brief explanation of pre-constructed NEN datasets from the bioinformatics domain is given in this section.
The structure for our proposed model is described in Section \ref{proposed_method}.
Experiment settings for testing model performances are provided in Section \ref{experiments}.
Furthermore, \ref{dataset} in Experiment Settings(subsection \ref{experiments}) provides the overview of preprocessing for data and financial NEN dataset construction with examples.
In Section 5, we present the details regarding the qualitative and quantitative analyses we conducted on the models.
Finally, in Section 6, we present our conclusions.

\section{Related Work}
\label{related_work}
Bioinformatics, chemical engineering, and materials science domain actively adopt cutting-edge deep learning frameworks for NEN tasks.
According to Cho et al. \cite{cho2017method}, various products exist for recognizing and normalizing named entities in biomedical fields such as ProMiner \cite{hanisch2005prominer} and MetaMap \cite{aronson2001effective}.
DNorm \cite{leaman2013dnorm} and TaggerOne \cite{leaman2016taggerone} also used machine learning models such as pairwise ranking scoring and semi-Markov models, respectively, for NEN processing.
In genetic engineering, GenNorm \cite{wei2011cross} and GNAT \cite{hakenberg2011gnat} are used to normalize the gene names.
ChemSpot \cite{rocktaschel2012chemspot} uses Conditional Random Field for NER and NEN tasks in chemical engineering.
Weston et al. \cite{weston2019named} developed MatScholar \cite{weston2019named} python repository to perform general NLP tasks on material science texts, which includes entity normalization.

The above researches and products used NEN datasets concentrated on specific domains.
ShARe/CLEF \cite{suominen2013overview} is one of the widely used NEN datasets for bioinformatics that is made up of clinical notes.
The NCBI \cite{dougan2014ncbi} dataset contains PubMed abstracts for disease name normalization tasks.
TAC2017ADR \cite{demner2018dataset} aims to link identical drug labels.
The BC2GM \cite{smith2008overview}, BioNLP09 \cite{kim2009overview}, and BioNLP-OST19 \cite{bossy2019bacteria} datasets deal with genes, proteins, and bacteria, respectively.
In chemical engineering, SCAI \cite{kolarik2008chemical} and IUPAC \cite{klinger2008detection} are available for researches on chemical name matching.
Similar to chemical names, Weston et al. \cite{weston2019named} developed a dataset for material engineering to normalize entities to a canonical form.

Applying machine learning algorithms in the financial domain is gaining increasing attention.
One major branch is stock movement forecasting using various deep learning mechanisms \cite{arratia2019evaluation, corba2020ar}.
Thanks to the rapid developments of unstructured data processing techniques, researches on applying text mining techniques to the financial fields have increased in number.
In their study, Gupta et al. \cite{gupta2020comprehensive} illustrated the trends for applying text mining in finance.
Among many related text mining applications in finance, NEN can be applied to various financial researches and financial practices.
In preprocessing for applying text mining techniques to solve real-world problems, NER and NEN models are performed preemptively.
However, the NEN dataset for the financial domain is scant and there is a need for developing a dataset targeting the financial NEN.

% Todo
Many researchers have developed targeted datasets for more general NEN tasks in domains such as user comments, product description, and financial invoices.
For example, in their study, Jijkoun et al. \cite{jijkoun2008named} used user comments from newspaper websites.
Sun et al. \cite{sun2012product} performed normalization of product entity names, for which the dataset was developed by the authors.
The study conducted by Francis et al. \cite{francis2019transfer} on financial invoices is the most relevant one to our study.
However, Francis et al. focused on insurance, telecommunications, banking, and tax companies using the following entities: International Bank Account Number (IBAN) of the beneficiary, invoice number, invoice date, and due date \cite{francis2019transfer}.
The focus of our study is on more general financial entity normalization, which covers entities from all financial sectors.
Previous studies using the datasets illustrated above used various machine learning and deep learning models.

There are similarities between the string matching methodologies in various other fields and NEN researches.
Sun et al. \cite{sun2012product} proposed NEN for product names using a pre-constructed product entity linkage dictionary.
In semantic string matching, Siamese Neural Networks are widely used  \cite{mueller2016siamese, ranasinghe2019semantic, liu2018matching}.
Krivosheev et al. \cite{krivosheev2020siamese} used Siamese Graph Neural Network for company name normalization.
We need to extend NEN on company names to NEN on a wide range of product names and legal entities.
Siamese RNN model successfully apprehends the morphological similarity between strings \cite{neculoiu2016learning}.
Niu et al. \cite{niu2019multi} applied Attention mechanisms for medical concept normalization.
Furthermore, the evolution of Transformer-based models capacitate the adoption pre-trained language models such as BERT \cite{devlin2018bert} for entity linking problems \cite{mulang2020evaluating}.

The major development in recent NEN researches is as follows.
D`Souza et al. \cite{d2015sieve} proposed an early NEN model using a rule-based model, which requires comparatively more human input when generating the rules.
The model is static and, thus, there is a possibility that new rules need to be created when applying the model to other datasets.
NEN models that use more advanced machine learning and deep learning techniques can be more effective.
Leaman et al. \cite{leaman2016taggerone} used semi-Markov model, Li et al. \cite{li2017cnn} used word-level CNN model, and Wirght and Dustin \cite{wright2019normco} and Phan et al. \cite{phan2019robust} models based on BiGRU and BiLSTM.
However, BERT achieved state-of-the-art performance in many general text mining and natural language processing (NLP) challenges.
Compared with the four models illustrated above, the most recent researches such as the BERT ranking model \cite{ji2020bert} and BioSyn \cite{sung2020biomedical} takes full advantage of the BERT model by training the model based on BERT embeddings.
The BERT Ranking model \cite{ji2020bert} used ranking-based objective function and BioSyn \cite{sung2020biomedical} used Synonym Marginalization techniques as the objective function for training.
Our proposed model optimizes BERT embedding vectors with named entity graph's edge weight updating neural network.
Our proposed model successfully captures the ground truth linkage between named entity graphs, achieving the highest accuracies.
Previous NEN researches focus mainly on the NEN dataset from a specific domain.
To test the efficacy of our model in more general NEN tasks, we evaluate our model with NEN datasets from both the bioinformatics domain and financial domain.

Many NEN researches explore semi-supervised learning models.
Our proposed model is motivated by one of the leading semi-supervised models on images, Edge-Labeling Graph Neural Network for Few-shot Learning \cite{kim2019edge} (EGNN).
The major difference between EGNN and our model is that EGNN labels an edge for each round of training but our model updates edge weights for top K connected entities.
By capturing more node and edge information simultaneously for each round of training, the proposed model shows better performance compared with other NEN models.

\section{Proposed Method}
\label{proposed_method}
Our proposed model, Edge Weight Updating Neural Network, consists of four major parts.
\begin{enumerate}
	\item Ground Truth Entity Graph construction
	\item Similarity-Based Entity Graph construction
	\item Edge Weight Updating Neural Network training
	\item Edge Weight Updating Neural Network inferencing
\end{enumerate}
The basic idea behind Edge Weight Updating Neural Network is to minimize the Ground Truth Entity Graph's edge weight distributions and the Similarity-Based Entity Graph's edge weight distributions.
Entity embeddings are trained with Kullback-Leibler divergence  \cite{kullback1951information} loss between two graphs.
Detailed steps for constructing the Ground Truth Named Entity Graph, building the Similarity-Based Entity Graph, and training and inferencing the Edge Weight Updating Neural Network are presented in Sections \ref{gteg}, \ref{sbeg}, \ref{ewunnt} and \ref{ewunni}, respectively.

\subsection{Ground Truth Entity Graph Construction}
\label{gteg}
\begin{figure}
	\includegraphics[width=1\textwidth]{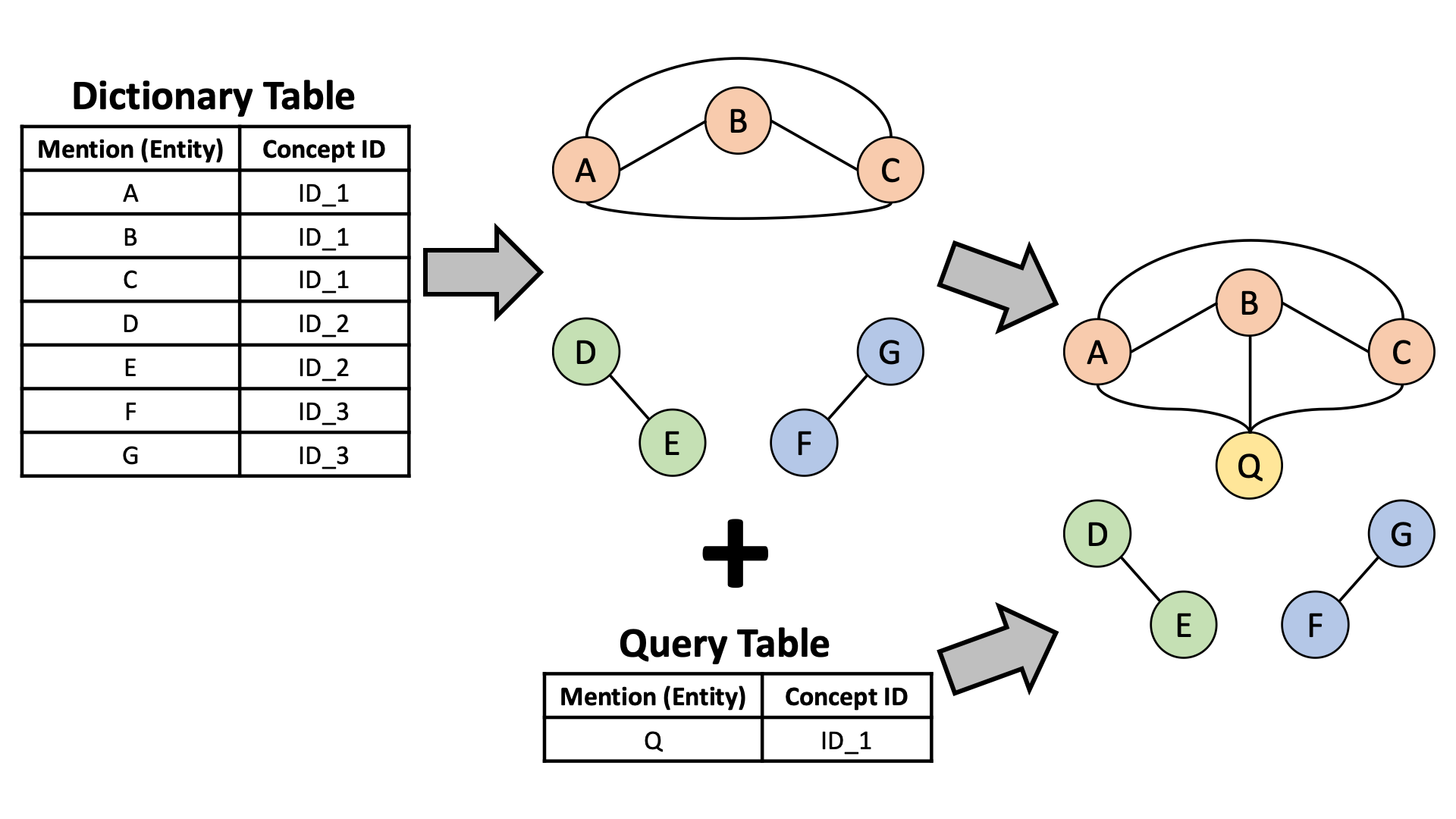}
	\caption{Ground Truth Named Entity Graph construction}
	\label{fig:ref_graph_construction}       % Give a unique label
\end{figure}
Ground Truth Entity Graph constructions are based on mentions (entities) in each dataset and their concept IDs.
Figure \ref{fig:ref_graph_construction} demonstrates the steps for building the graph.

For the NEN corpus, each entity is annotated with one or more concept IDs.
For example in Figure \ref{fig:ref_graph_construction}, entities A, B, and C share the same concept ID, ID\_1.
Then, entities A, B, and C are fully connected in the entity graph.
Other entity pairs, D - E (concept ID: ID\_2) and  F - G (concept ID: ID\_3) are linked.
The training dataset for each NEN corpus has query entities with corresponding concept ID.
If query entity Q has a concept ID of ID\_1, then, query entity Q will be linked to entities A, B, and C in the pre-constructed graph.
As the constructed graph is the ground truth graph, each edge weight in the graph is 1.

We iterate all the entities in training sets that include the referencing dictionary entity table and the query entity table.
Graph created by the following steps above is the Ground Truth Entity Graph which is the reference or the target graph the Similarity-Based Entity Graph will try to match.

\subsection{Similarity-Based Entity Graph Construction}
\label{sbeg}
\begin{figure}
	\includegraphics[width=1\textwidth]{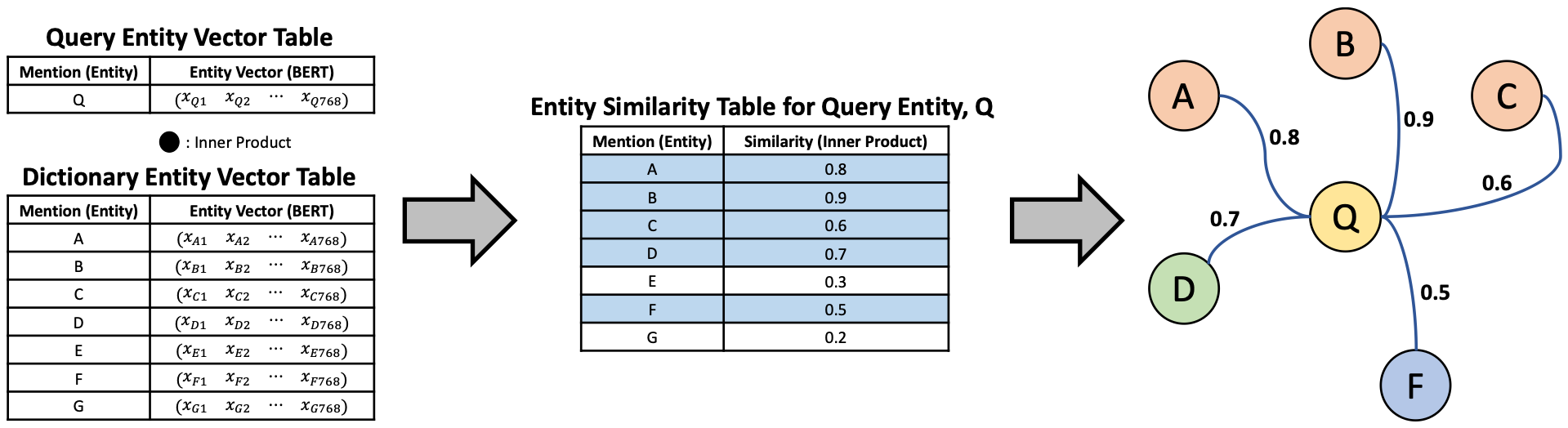}
	\caption{Entity Matching Graph based on Entity Similarity construction}
	\label{fig:sim_graph_construction}       % Give a unique label
\end{figure}
For each query entity, Similarity-Based Entity Graph is constructed as follows.
Graph edges are calculated using BERT embedding vector similarities.
We use BioBERT \cite{lee2020biobert} for bioinformatics NEN corpus' initial BERT embeddings and the original BERT \cite{devlin2018bert} for financial NEN corpus' initial BERT embeddings.

For example, in Figure \ref{fig:sim_graph_construction}, let query entity Q has size of 768 (vector length of BERT embeddings), $Embed_Q = (X_{Q1}~~X_{Q2}~~\cdots~~X_{Q768})$.
Similarly, BERT-based entity embeddings in the dictionary set are also denoted as $Embed_{entity} = (X_{entity1}~~X_{entity2}~~\cdots~~X_{entity768})$.
The BERT embedding has a fixed length of 768, so our embedding vectors have a vector length of 768.

To calculate the edge weights based on entity similarities, we calculate inner products between query entities and dictionary entities.
$<~,~>$ is the notation for inner product and $Sim_Q$ is the set of similarities between query entity Q and all the entities in a dictionary; then the similarity between each query entity and each dictionary entity calculation is expressed as Equation \ref{eq:sim_calc},
\begin{equation}
	\label{eq:sim_calc}
	\begin{aligned}
		 &
		Sim_Q = \{Sim | Sim = <Embed_Q,~Embed_D>~for~D \in Dictionary Entity Set\} \\
		 &
		where,                                                                     \\
		 &
		Embed_{Q \in Query Entity Set} = (X_{Q1}~~X_{Q2}~~\cdots~~X_{Q768}),       \\
		 &
		Embed_{D \in Dictionary Entity Set} = (X_{D1}~~X_{D2}~~\cdots~~X_{D768})   \\
	\end{aligned}
\end{equation}

We normalize the similarity score by dividing the maximum similarity score in each query entity's similarity score set, $Sim_Q$.
For Similarity-Based Entity Graph, top K edges based on similarity score are selected.
Highlighted blue region in entity similarity table for query entity Q in Figure \ref{fig:sim_graph_construction} demonstrates the edge weight determination steps when $K=5$.
Mathematically, edge weights are calculated using Equation \ref{eq:weight_calc}.
\begin{equation}
	\label{eq:weight_calc}
	\begin{aligned}
		 &
		ConnectedEdgeWeight = \{Weight_{Edge} | Edge \in ConnectedEdge\}                  \\
		 &
		where,                                                                            \\
		 &
		ConnectedEdge = argmax_k\{Weight_Q\},                                             \\
		 &
		Weight_Q = \left\{Weight | Weight = \frac{Sim}{Max_Q}~for~ Sim \in Sim_Q\right\}, \\
		 &
		and, Max_Q = max\{Sim_Q\}                                                         \\
	\end{aligned}
\end{equation}

For each training epoch, which is illustrated in Section \ref{ewunnt}, edge weights are updated.
Updated entity embedding vectors generate new similarity scores that alter the edge weights in the graph.

\subsection{Edge Weight Updating Neural Network Training}
\label{ewunnt}
\begin{figure}
	\includegraphics[width=1\textwidth]{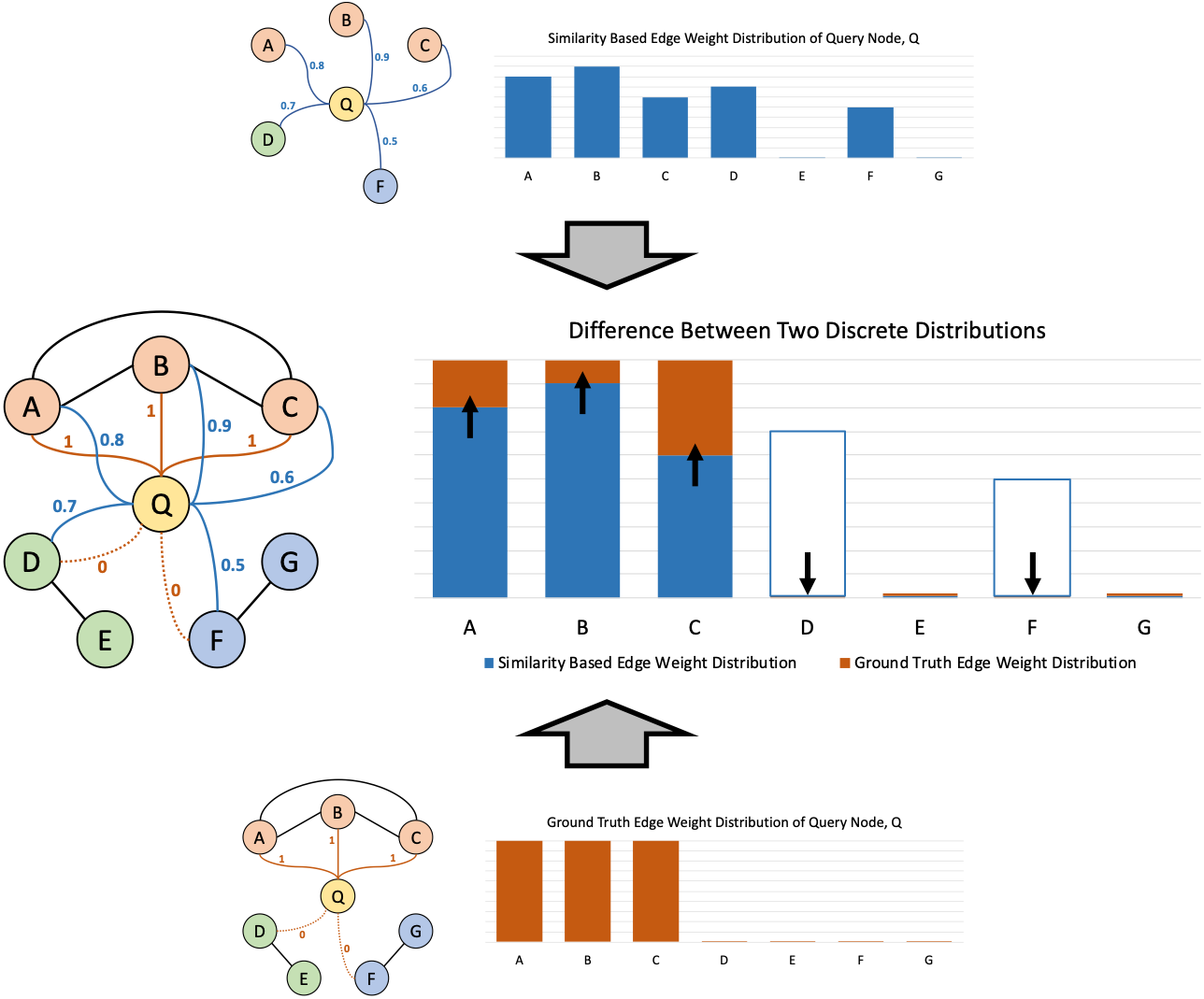}
	\caption{Minimizing the Edge Weight Distributions in Edge Weight Updating Neural Network for Query Entity Q}
	\label{fig:kldiv}       % Give a unique label
\end{figure}
The main concept of Edge Weight Updating Neural Network is to minimize the difference between the edge weights' discrete distribution for each query entity in the Ground Truth Entity Graph and the Similarity-Based Entity Graph.
As illustrated in Section \ref{sbeg}, edge weights are calculated by entities' embeddings.
In each training epoch in Edge Weight Updating Neural Network, baseline BERT model's parameters are optimized to mimic the ground truth edge weight distributions.

Figure \ref{fig:kldiv} shows the training process of our proposed model for the number of connected edges in the Similarity-Based Entity Graph is 5 ($K=5$).
Following the example in Section \ref{sbeg}, query entity Q is connected to dictionary entities A, B, C, D, and F, and edge weights are 0.8, 0.9, 0.6, 0.7, and 0.5, respectively.
Given the Ground Truth Entity Graph in Section \ref{gteg}, the truth edge weights for connected edges between query entity Q and dictionary entities, A, B, C, D, and F are 1, 1, 1, 0, and 0, respectively.

In training procedures, BERT parameters are tuned to make edge weights distributions in Similarity-Based Entity Graph closer to the ground truth edge weight distributions.
We use Kullback-Leibler Divergence Loss \cite{kullback1951information}(KL divergence loss, henceforth) for training our model.
As edge weight distribution is discrete, we normalize the edge weights using the Softmax function.

We denote graph as $G$, entity as $V$, and edge as $E$.
The Ground Truth Entity Graph and the Similarity-Based Entity Graph are denoted as $G_{GT} = (V_{GT},E_{GT})$ and $G_{Sim}=(V_{Sim},E_{Sim})$, respectively.
The adjacency matrices for Ground Truth Entity Graph and the Similarity-Based Entity Graph are denoted $GT\_A$ and $Sim\_A$.
$P_{Sim\_Edge_Q}$ is the discrete distribution of edge weights of Q in Similarity-Based Entity Graph.
$P_{GT\_Edge_Q}$ is the discrete distribution of edge weights of Q in the Ground Truth Entity Graph.
Our KL divergence loss is calculated using Equation \ref{eq:kldiv}.
\begin{equation}
	\label{eq:kldiv}
	\begin{aligned}
		 &
		Loss=KL(P_{GT\_Edge_Q}||P_{Sim\_Edge_Q})=P_{GT\_Edge_Q} \cdot \log \left( \frac{P_{GT\_Edge_Q}}{P_{Sim\_Edge_Q}} \right) \\
		 &
		where,                                                                                                                   \\
		 &
		P_{GT\_Edge_Q}=Softmax(GT\_Edge_Q),                                                                                      \\
		 &
		P_{Sim\_Edge_Q}=Softmax(Sim\_Edge_Q),                                                                                    \\
		 &
		GT\_Edge_Q=\{GT\_A_{query,d}|d \in argmax_k\{Sim\_A_Q\}\},                                                               \\
		 &
		Sim\_Edge_Q=\{Sim\_A_{query,d}|d \in argmax_k\{Sim\_A_Q\}\},                                                             \\
		 &
		and, A_Q~is~the~edge~weight~vector~connected~to~Q~for~given~query~entity~node~Q                                          \\
	\end{aligned}
\end{equation}
We use an Adam optimizer with weight decay \cite{loshchilov2017decoupled}, and set the batch size to 16 and the number of connected edges in the Similarity-Based Entity Graph to 30 ($K=30$) for all datasets we test.
We train our model for 50 epochs.
The best scores are reported in Section \ref{experiments}.

\subsection{Edge Weight Updating Neural Network Inferencing}
\label{ewunni}
\begin{figure}
	\includegraphics[width=1\textwidth]{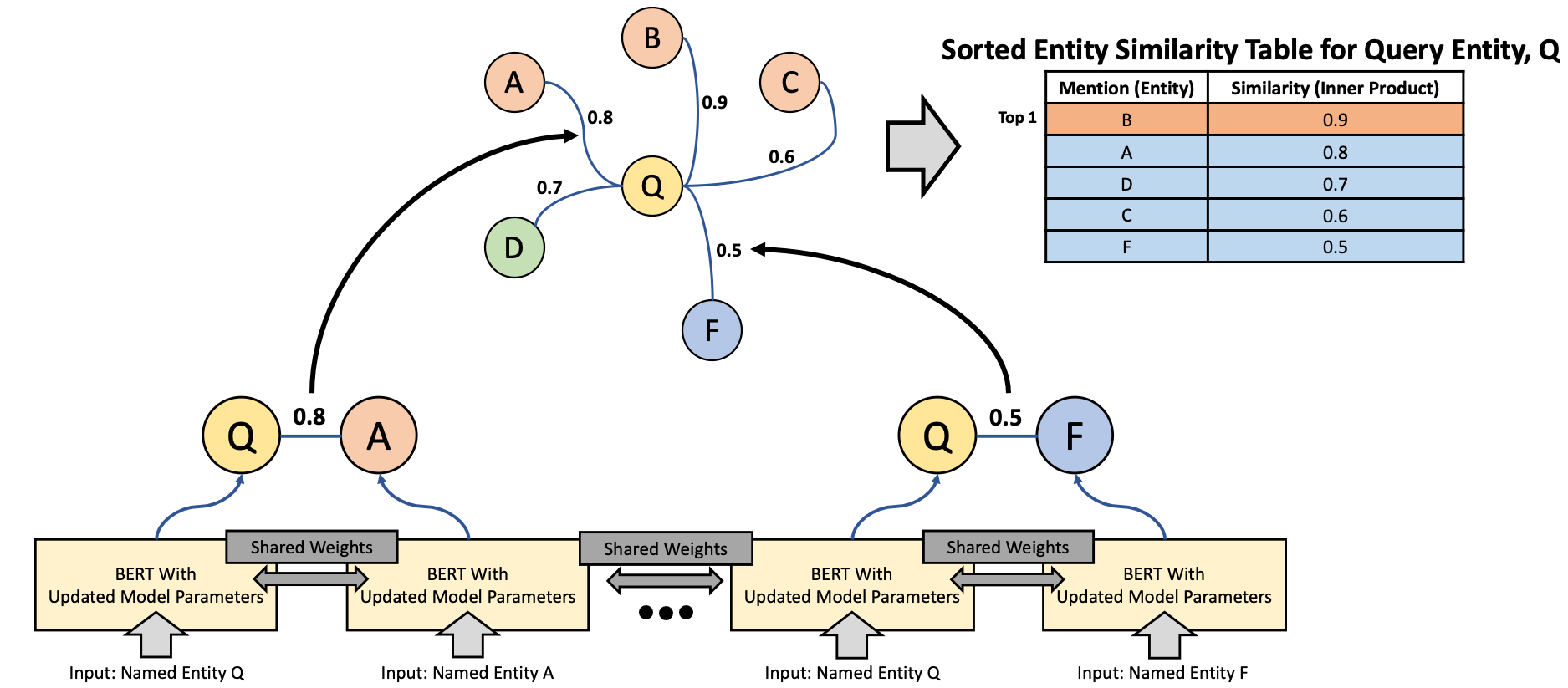}
	\caption{Inferencing the Edge Weight Distributions in Edge Weight Updating Neural Network for Query Entity Q}
	\label{fig:inference}       % Give a unique label
\end{figure}
First, fine-tuned BERT embeddings illustrated in Section \ref{ewunnt} are used to embed unseen query entities in test sets.
With newly computed BERT embedding vectors, we repeat the steps in Section \ref{sbeg} to construct the new Similarity-Based Entity Graph.
For each query entity, a dictionary entity with the highest edge weights is returned as a synonym.
Figure \ref{fig:inference} demonstrates the inferencing process of the Edge Weight Updating Neural Network.

\section{Experiment Settings}
\label{experiments}
\subsection{Dataset}
\label{dataset}
\subsubsection{Named Entity Normalization Datasets in Bioinformatics}
\label{NEN_bio}
Most NEN researches are from the bioinformatics domain.
To test our model's performance with other NEN models, we select three of the most used bioinforinmatics NEN datasets: NCBI Disease \cite{dougan2014ncbi} and two datasets from Biocreinative V CDR (BC5CDR, henceforth) \cite{li2016biocreative}.

Three datasets summarized below contains bioinformatics-related mentions (entities) with unique concept IDs.
The main goal of these datasets is to identify the mentions that share the same concept IDs.
We follow NEN preprocessing convention for the datasets below, in which the mentions that do not exist in the concept dictionary are eliminated \cite{phan2019robust}.
Bioinformatics NEN datasets usually consist of train, development, and test sets.
Following previous studies, we use train and development sets for training our model.
Test sets are used for evaluations.

\begin{table}
	\caption{Data Statistics of Three Bioinformatics NEN Datasets}
	\label{dataset_bio}
	\centering
	\begin{tabular}{lllllll}
		\hline\noalign{\smallskip}
		                & \multicolumn{3}{l}{\# of Documents} & \multicolumn{3}{l}{\# of Mentions (Entities)}                                \\
		                & Train                               & Dev                                           & Test & Train & Dev   & Test  \\
		\noalign{\smallskip}\hline\noalign{\smallskip}
		NCBI Disease    & 592                                 & 100                                           & 100  & 5,134 & 787   & 960   \\
		BC5CDR Disease  & 500                                 & 500                                           & 500  & 4,182 & 4,244 & 4,424 \\
		BC5CDR Chemical & 500                                 & 500                                           & 500  & 5,203 & 5,347 & 5,385 \\
		\noalign{\smallskip}\hline
	\end{tabular}
\end{table}

\paragraph{NCBI Disease} \cite{dougan2014ncbi}.
NCBI Disease corpus provides disease mentions in different surface forms.
Disease mentions in this dataset are extracted from 793 PubMed abstracts containing a total of 6,892 disease mentions, which are mapped to 790 unique disease concepts.
Disease concepts are annotated by Medical Subject Headings (MeSH) and Online Mendelian Inheritance in Man (OMIM).
Disease mentions sharing the same disease concept are considered synonyms.
Table \ref{dataset_bio} shows detailed statistics of the NCBI Disease corpus.

\paragraph{Biocreative V CDR Disease and Biocreative V CDR Chemical} \cite{li2016biocreative}.
The BC5CDR corpus is organized for challenging tasks of disease named entity recognition and chemical-induced disease relation extraction.
The BC5CDR corpus consists of 1,500 PubMed articles with 4,409 annotated chemicals, and 5,818 disease and 3,116 chemical-disease interactions \cite{li2016biocreative}.
The dataset contains disease mention corpus and chemical mention corpus.
Disease mentions are mapped into the MeSH IDs similar to the NCBI Disease corpus.
Chemical mentions are annotated using the Comparative Toxicogenomics Database (CTD) \cite{davis2009comparative}.
Mentions that share the same disease concept and chemical concept based on MeSH ID and CTD ID are considered synonyms.
Detailed statistics of both BC5CDR Disease corpus and BC5CDR Chemical corpus are illustrated in Table \ref{dataset_bio}.

\subsubsection{Named Entity Normalization Datasets in Finance}
\label{NEN_finance}
There are no publicly open financial NEN datasets available; therefore, we constructed our own financial NEN dataset to test the performance of our proposed model in NEN tasks other than the bioinformatics domain.

\begin{figure}
	\includegraphics[width=1\textwidth]{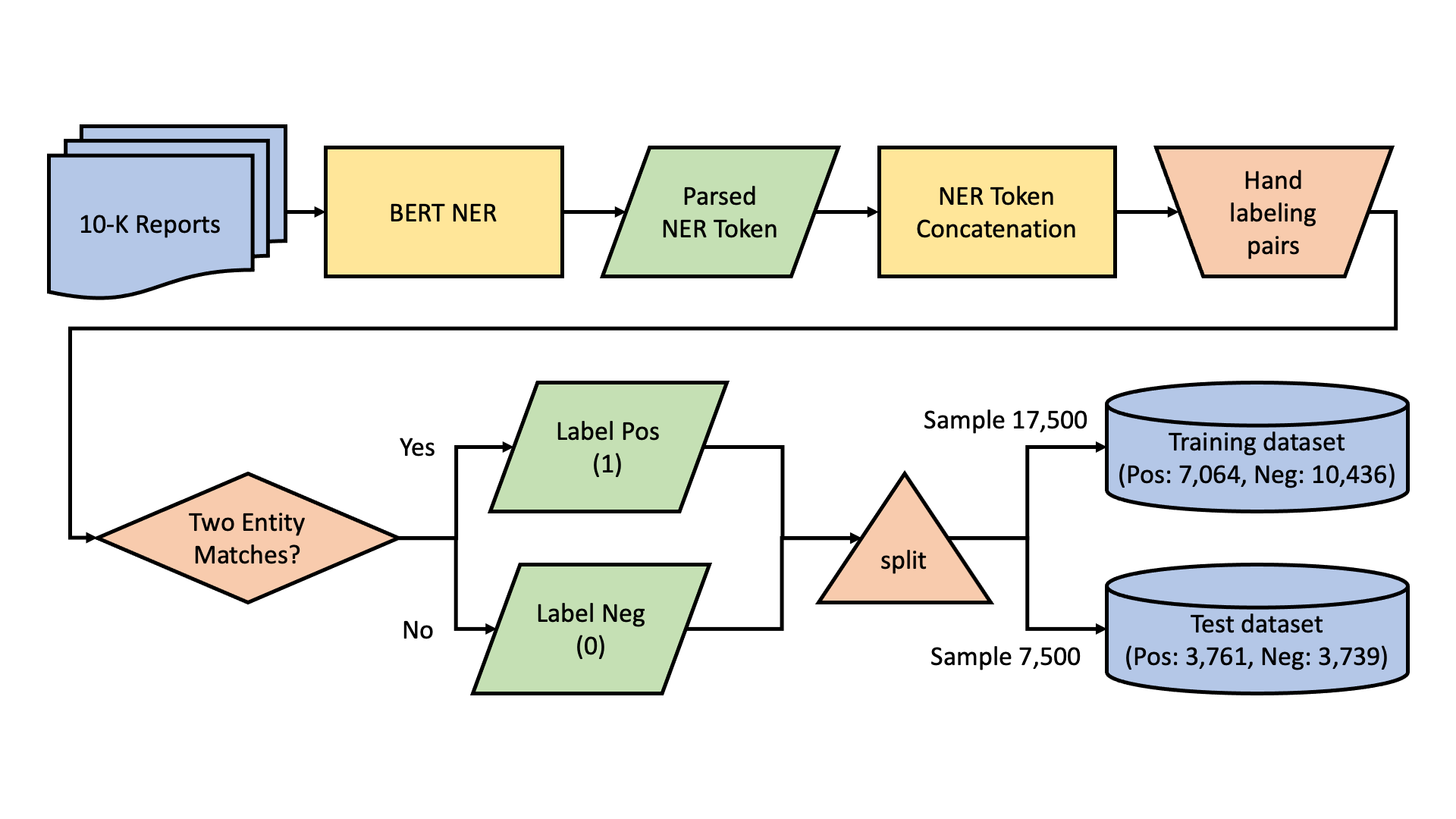}
	\caption{Flow Diagram of the Overall Dataset construction}
	\label{fig:flowdiagram}       % Give a unique label
\end{figure}

\paragraph{Overview}.
\label{dataset_overview}
We construct the dataset for the financial NEN task from the annual reports (Form 10-K) of Standard and Poor's 500 listed companies.
We aim to build the dataset that fulfills the need for financial NEN; the dataset includes (1) synonyms, (2) abbreviations, (3) acronyms, (4) different combinations of punctuations and alphabets, (5) descriptive phrases, and (6) possible NER parsing errors.
A detailed explanation of primary data sources, data preprocessing steps, and dataset construction procedures are as follows.
Fig.~\ref{fig:flowdiagram} demonstrates the overall flow diagram for NEN dataset construction.

\paragraph{Data Source}.
We gather the year 2019's Form 10-Ks (published early 2020) of S\&P500 companies from the U.S. firms and Exchange Commission (SEC) website\footnote{https://www.sec.gov/edgar.shtml}, which is open to the public.
We parse the business section of each 10-K documents from 496 companies.
The business section of 10-K is considered the self-identity of firms and presents the information of main products, competitors, partners, and laws affecting the business.
Among the sections in 10-K, this section contains the most number of entities.
Out of 496 companies' business section, 67,792 sentences were parsed.

\paragraph{Data Preprocessing}.
\label{data_preprocessing}
For NER in financial documents, we implement the BERT NER model \cite{devlin2018bert} using Huggingface's\footnote{https://huggingface.co} Python repository.
Huggingface's NER model is trained using CoNLL-2003 NER dataset \cite{sang2003introduction}.
The outputs of the BERT NER model are WordPiece tokens that we have to link together with specified rules that will be circumstantially described below.
There are four types of entity types: persons (PER), organizations (ORG), locations (LOC) and miscellaneous names (MISC), and one outside the named entity tag (O) in the CoNLL-2003 dataset.
We detect entities with ORG and MISC tags.
For the year 2019 S\&P500 firms' 10-K, we parse a total of 41,593 named entities.

\paragraph{Dataset Construction}.
\begin{table}
	\caption{Example of Financial Named Entity Normalization Dataset}
	\label{dataset_example}
	\centering
	\begin{tabular}{lllll}
			\hline\noalign{\smallskip}
			                             & Named Entity                                      & Matching Named Entity                            \\
			\noalign{\smallskip}\hline\noalign{\smallskip}
			Synonyms                     & Coca-Cola \textsuperscript{\textregistered}       & Coca-Cola                                        \\
			                             & COVID-19 Pandemic                                 & COVID-19                                         \\
			                             & iPhone 11 Pro Max                                 & iPhone \textsuperscript{\textregistered }        \\
			Abbreviations                & Baker Hughes Company                              & Baker Hughes Co.                                 \\
			                             & Comcast Corporation                               & Comcast Corp.                                    \\
			                             & Qualcomm Incorporated                             & Qualcomm Inc.                                    \\
			Acronyms                     & Amazon Web Services                               & AWS                                              \\
			                             & Bank of New York Mellon                           & BNY Mellon                                       \\
			                             & New York Stock Exchange                           & NYSE                                             \\
			Combinations of punctuations & Apple, Inc.                                       & Apple Inc.                                       \\
			                             & Walmart U. S.                                     & Walmart U. S                                     \\
			                             & Booz Allen \& Hamilton                            & Booz Allen Hamilton                              \\
			Descriptive Phrases          & EY ( formerly Ernst \& Young )                    & Ernst and Young                                  \\
			                             & Securities Exchange Act of 1934                   & ( the Exchange Act )                             \\
			                             & Facebook (including Instagram)                    & Facebook \textsuperscript{\textregistered}       \\
			NER Parsing Errors           & Disney Channel-the                                & Disney Channel                                   \\
			                             & Full Throttle \textsuperscript{\textregistered}-a & Full Throttle \textsuperscript{\textregistered}) \\
			                             & Keystone-our                                      & Keystone Foods                                   \\
			\noalign{\smallskip}\hline
		\end{tabular}
\end{table}

With named entities recognized illustrated in Section \ref{data_preprocessing}, we construct the financial named entity normalization dataset.
As mentioned in Section \ref{dataset_overview}, our focus is to build a NEN dataset to meet the need for general text mining in finance; the dataset includes (1) synonyms, (2) abbreviations, (3) acronyms, (4) different combinations of punctuations and alphabets, (5) descriptive phrases, and (6) possible NER parsing errors.
We hand label a total of 7,155 unique named entities into 2,600 groups; with each group sharing the same identity.
Table \ref{dataset_example} shows three examples in our dataset for types of named entities that need to be normalized.
\begin{itemize}
	\item Synonyms:

	      There exist entities with the suffix ``\textsuperscript{\textregistered}" or ``\texttrademark".
	      ``Coca-Cola \textsuperscript{\textregistered}" and ``Coca-Cola" are the same entity.
	      In addition, ``COVID-19 Pandemic" and ``COVID-19" should be linked.
	      We generalize the product model numbers in which ``iPhone 11 Pro Max" and ``iPhone \textsuperscript{\textregistered}" are considered identical entities.

	\item Abbreviations:

	      Most abbreviations occur for abridging ``Company" to ``Co.", ``Corporation" to ``Corp.", and ``Incorporated" to ``Inc.".

	\item Acronyms:

	      Acronyms are one of the most challenging NEN tasks.
	      There are multiple abbreviations that are included in financial documents.
	      We avoided matching acronyms if there are multiple original entities can be assigned.
	      For example, ``Advanced Development Programs ( ADP )" and ``Automatic Data Processing, Inc. ( ADP )" both share the same acronyms, ``ADP", but these should not be linked together.

	\item Combinations of punctuations:

	      The different combinations of punctuations problems can be solved using rule-based approaches.
	      However, there are many entities with a combination of punctuations.
	      ``,", ``.", and ``\&" are commonly found and used interchangeably.

	\item Descriptive phrases:

	      In parsed named entity, an entity with descriptive phrases can be frequently found.
	      With or without descriptive phrases, the root or the identified entity is invariable.

	\item NER parsing errors:

	      No NER models and entity concatenation models are perfect.
	      If NER is conducted manually, there are possible human errors too.
	      According to our dataset, one common error model makes is appending the following token after ``-" token.
	      NER parsing error correction is one of the important targets our NEN model aims to achieve.
\end{itemize}

\begin{table}
	\caption{Statistics of the Financial Named Entity Normalization Dataset}
	\label{dataset_stat}
	\centering
	\begin{tabular}{lllll}
		\hline\noalign{\smallskip}
		                              & Train  & Test  & Total  \\
		\noalign{\smallskip}\hline\noalign{\smallskip}
		\# of Identical Entity Groups & 2,510  & 90    & 2,600  \\
		\# of Positive Pairs          & 7,064  & 3,761 & 10,825 \\
		\# of Negative Pairs          & 10,436 & 3,739 & 14,175 \\
		\# of Pairs Total             & 17,500 & 7,500 & 25,000 \\
		\noalign{\smallskip}\hline
	\end{tabular}
\end{table}

Hand-matched entity pairs are labeled positive.
We also added negatively labeled pairs in which two entities have no relationship.
A total of 25,000 pairs with 10,825 positive matching pairs and 14,175 negative pairs are created.
We separate entity groups for a train set and test set in which there are no overlapping groups.
This eliminates possible training bias, especially when training the model with entities' graph topology.
Table \ref{dataset_stat} shows the statistics of our financial NEN dataset.

\subsection{Experiment Settings: Named Entity Normalization in Bioinformatics}
\label{experiments_bio}
We compare our proposed model's performance with seven different biomedical NEN models.
The accuracy score presented in this study is excerpted from original papers.
A summary of each model is illustrated in Table \ref{bioinfo_model}.

\begin{table}
	\caption{Models Used in Bioinformatics NEN datasets evaluations}
	\label{bioinfo_model}
	\centering
	\begin{tabular}{p{0.22\textwidth}p{0.68\textwidth}}
		\hline\noalign{\smallskip}
		Models                                & Descriptions                                                                                                                                                                                                         \\
		\noalign{\smallskip}\hline\noalign{\smallskip}
		Sieve-based  \cite{d2015sieve}        & This is one the earliest NEN papers.
		The research was conducted with 10 Sieve, which is mostly a rule-based approaches.
		Many published post this research follow similar preprocessing steps.                                                                                                                                                                                        \\

		\noalign{\smallskip}
		Taggerone  \cite{leaman2016taggerone} & Taggerone used the semi-Markov model for both NER and NEN tasks.
		Taggerone was originally validated on the NCBI Disease and BC5CDR corpus.                                                                                                                                                                                    \\

		\noalign{\smallskip}
		CNN Ranking  \cite{li2017cnn}         & CNN Ranking model used a word-level deep learning approach for NEN.
		This research did not perform better than the previous model, Taggerone.
		However, it was the first study that applied deep learning to NEN tasks.                                                                                                                                                                                     \\

		\noalign{\smallskip}
		NormCo  \cite{wright2019normco}       & NormCo used BiGRU, which is considered to be a better performing deep learning model with text data.
		NormCo achieved similar accuracy scores with significantly fewer parameters.                                                                                                                                                                                 \\

		\noalign{\smallskip}
		BNE  \cite{phan2019robust}            & BNE introduced two-level BiLSTM to capture both character-level and word-level information of biomedical entities, achieving increased NEN performance.                                                              \\

		\noalign{\smallskip}
		BERT Ranking  \cite{ji2020bert}       & BERT Ranking model is based on Transformer-based embeddings that use the pre-trained BERT \cite{devlin2018bert}, BioBERT \cite{lee2020biobert}, and ClinicalBERT \cite{si2019enhancing} for their entity embeddings.
		For each entity, candidate concepts were retrieved and three different BERT models are fine-tuned to rank and to capture the ground truth concepts.                                                                                                          \\

		\noalign{\smallskip}
		TripletNet  \cite{mondal2020medical}  & The concept of TripletNet \cite{hoffer2015deep} for semi-supervised learning was introduced for NEN tasks.
		This study uses CNN for entity embedding and shared CNN parameters are trained with TripletNet structure.                                                                                                                                                    \\

		\noalign{\smallskip}
		BioSyn  \cite{sung2020biomedical}     & BioSyn uses BioBERT for entity embeddings and trained with Synonym Marginalization.
		Marginal Maximum Likelihood (MML) is the objective function for Synonym Marginalization.                                                                                                                                                                     \\
		\noalign{\smallskip}\hline
	\end{tabular}
\end{table}

\subsection{Experiment Settings: Named Entity Normalization in Finance}
\label{experiments_finance}
% As mentioned in Section \ref{proposed_method}, we jointly train Siamese BiLSTM and Siamese GCN.
The dataset we used is covered in Section \ref{NEN_finance}.
% There is no Named Entity Normalization dataset in Finance, so we conduct traditional NEN methods to compare the performance.
Table \ref{finance_model} shows each model used in NEN in Finance is tested.
The experiments are conducted using Intel Core-i9-10940X CPU with 128GB memory and three NVIDIA GeForce Titan RTX GPU.
To avoid possible biases caused by exogenous variables, we use the same setting for all models if applicable.

\begin{table}
	\caption{Models Used in Finance NEN datasets evaluations}
	\label{finance_model}
	\centering
	\begin{tabular}{p{0.25\textwidth}p{0.65\textwidth}}
		\hline\noalign{\smallskip}
		Models                                          & Descriptions                                                                                         \\
		\noalign{\smallskip}\hline\noalign{\smallskip}
		Edit Distance \cite{levenshtein1966binary}      &
		Edit Distance is suitable for basic NEN tasks for linking ``Apple Inc" and ``Apple Inc.".
		However, Edit Distance can only capture the superficial morphological similarity between two entities.
		In our experiment, we calculate the Edit Distance between two entity pairs and train a simple classifier to determine the equivalence of two entities. \\
		\noalign{\smallskip}
		BERT \cite{devlin2018bert}                      &
		BERT is a state-of-the-art model for various NLP tasks.
		However, for our specific tasks, the BERT model has a limitation on capturing morphological similarity between entity pairs.
		We use pre-trained BERT vectors with size 768 and train a simple MLP classifier with batch size 4096 to determine the linkage between entity pairs.    \\
		\noalign{\smallskip}
		Siamese GCN \cite{kipf2016semi}                 &
		We use the entity graph illustrated in Section \ref{sbeg} and we use a pre-trained BERT vector for each entity node vector.
		2-layer Siamese GCN is used in our experiment with 256 hidden nodes for the first GCN layer and 16 hidden nodes for the second GCN layer.
		GCN requires more epochs for training so we trained for 120 epochs for the full dataset (full batch: 17,500 entity pairs).
		The learning rate for ADAM optimizer for GCN is 0.01.                                                                                                  \\
		\noalign{\smallskip}
		Siamese BiLSTM \cite{schuster1997bidirectional} &
		For Character Level Siamese BiLSTM model training, we one-hot encoded the characters entity strings with unique 85 tokens.
		We stack two BiLSTM layers.
		The BiLSTM cells in the first layer return 64 dimension hidden states output and the BiLSTM cells in the second layer return 16 dimension hidden states output.
		To prevent overfitting, we train the BiLSTM model for 12 epochs.
		The BiLSTM model is trained with a learning rate of 0.001.
		Embedding dimension, 16, is the same as GCN.                                                                                                           \\
		\noalign{\smallskip}\hline
	\end{tabular}
\end{table}

\section{Results}
\label{results}
We conduct both quantitative and qualitative analysis.
For NCBI Disease, BC5CDR Disease, and BC5CDR Chemical datasets, we compare our proposed model's score with previous researches.
Bioinformatics datasets are reported by top one recommendation accuracy.
Given the biomedical entity in the train set, entities are matched with the most similar entities in datasets.
If the query entity and target entity share the same concept ID, it is considered correct.
The financial NEN dataset is a pairwise NEN matching corpus.
For evaluations on the financial NEN dataset, models that are used in evaluations distinguish whether two named entity pairs share identical meanings or not.
We also perform the qualitative analysis to assess models' weaknesses.

\subsection{Quantitative Analysis: Bioinformatics}
\label{quantitative_analysis_bio}
\begin{table}
	\caption{Bioinformatics Named Entity Normalization Performance Test}
	\label{result_table_bio}       % Give a unique label
	\centering
	\begin{tabular}{lllll}
			\hline\noalign{\smallskip}
			                                      & NCBI Disease  & BC5CDR Disease & BC5CDR Chemical \\
			\noalign{\smallskip}\hline\noalign{\smallskip}
			Sieve-Based  \cite{d2015sieve}        & 84.7          & 84.1           & 90.7            \\
			Taggerone  \cite{leaman2016taggerone} & 87.7          & 88.9           & 94.1            \\
			CNN Ranking  \cite{li2017cnn}         & 86.1          & -              & -               \\
			NormCo  \cite{wright2019normco}       & 87.8          & 88.0           & -               \\
			BNE  \cite{phan2019robust}            & 87.7          & 90.6           & 95.8            \\
			BERT Ranking  \cite{ji2020bert}       & 89.1          & -              & -               \\
			TripletNet  \cite{mondal2020medical}  & 90.0          & -              & -               \\
			BioSyn  \cite{sung2020biomedical}     & 91.1          & 93.2           & 96.6            \\
			\textbf{Proposed Model}               & \textbf{91.7} & \textbf{93.4}  & \textbf{96.7}   \\
			\noalign{\smallskip}\hline
		\end{tabular}
\end{table}

Table \ref{result_table_bio} shows a performance comparison between our proposed model and previous state-of-the-art models.
For three bioinformatics datasets, our proposed model achieved the highest accuracy.
Our model showed the highest performance increase by 0.6\% in the NCBI Disease corpus.
For BC5CDR Disease and BC5CDR Chemical corpus, the performance increase compared the previous state-of-the-art model is 0.2\% and 0.1\%, respectively.

The NCBI Disease corpus is a comparatively harder dataset based on the performance of other models.
We conclude that there there is a significant to increase the accuracy in a relatively lower performing dataset.
The previous model already performs excellently on the the BC5CDR corpus with an accuracy score increasing from 93.2\% to 96.6\%.
Significant performance increase in these datasets can be marginal.

\subsection{Quantitative Analysis: Finance}
\label{quantitative_analysis_finance}
\begin{table}
	\caption{Precision, Recall, F-Score, Accuracy of Models}
	\label{result_table_finance}       % Give a unique label
	\centering
	\begin{tabular}{lllll}
			\hline\noalign{\smallskip}
			                        & Precision (\%) & Recall (\%)    & F-Score (\%)   & Accuracy (\%)  \\
			\noalign{\smallskip}\hline\noalign{\smallskip}
			Edit Distance           & 43.12          & 63.35          & 51.31          & 62.60          \\
			BERT                    & 62.92          & 82.17          & 71.27          & 76.81          \\
			Siamese GCN             & 79.00          & 82.16          & 80.55          & 82.56          \\
			Siamese BiLSTM          & 75.96          & \textbf{89.98} & 82.38          & 85.15          \\
			\textbf{Proposed Model} & \textbf{93.43} & 82.11          & \textbf{87.40} & \textbf{88.13} \\
			\noalign{\smallskip}\hline
		\end{tabular}
\end{table}
Table \ref{result_table_finance} shows the performance of each model we test.
The evaluation metrics are expressed as follows
\begin{equation}
	\label{eq:metric}
	\begin{aligned}
		 &
		precision=\frac{tp}{tp+fp} \\
		 &
		recall=\frac{tp}{tp+fn}    \\
		 &
		F1=2\cdot \frac{precision\cdot recall}{precision+recall}
	\end{aligned}
\end{equation}

False positive indicates that two entities should not be matched, but our proposed model decided to link two entities.
False negative indicates that two entities should be matched, but our proposed model failed to link two entities.

For practical use in the NEN model in the finance domain, a model with higher precision should be rewarded more.
In practice, a model with higher precision will reduce the burden for practitioners' tasks by giving more reliable entity-matching results.
A model with higher precision will reduce time double-checking the validity entity pairs marked as matched.

Edit Distance had the lowest score along with all performance evaluation indicators.
Graph Convolutional Network we use for the experiments adopts the BERT vector as entity node features.
BERT and GCN have a similar recall, but GCN has higher precision, which brings higher F-score and accuracy compared with BERT.
Our proposed model achieved the highest precision, F-score, and accuracy.
Among all the models, our proposed model is the only model with a precision score over 90\%.
Therefore, our proposed model is the most suitable for practical use.

\subsection{Qualitative Analysis}
\label{qualitative_analysis}
\subsubsection{Error Analysis}
\label{error_analysis}
In error analysis, entities for which accurate recommendations are not made are reported.
Through error analysis, we aim to recognize the pattern of cases where recommendations are not properly made.

\begin{table}
	\caption{Error Analysis on Three Biomedical NEN datasets}
	\label{error_bio}
	\centering
	\begin{tabular}{lll}
			\hline\noalign{\smallskip}
			                & Query Entity             & Retrieved Synonym Entity        \\
			\noalign{\smallskip}\hline\noalign{\smallskip}
			                & encephalopathy           & aids encephalopathy             \\
			                & nail dystrophy           & twenty nail dystrophy           \\
			NCBI Disease    & cdm                      & cdmd                            \\
			                & copper overload          & copper deficient                \\
			                & g m2 gangliosidosis      & g m2 gangliosidosis type ii     \\
			\noalign{\smallskip}\hline\noalign{\smallskip}
			                & lung mass                & liver mass                      \\
			                & hypoactivity             & hyperactivity                   \\
			BC5CDR Disease  & htn                      & htx                             \\
			                & thrombocytopenia type ii & thrombocytopenia 2              \\
			                & chronic liver disease    & chronic hepatitis               \\
			\noalign{\smallskip}\hline\noalign{\smallskip}
			                & inorganic as             & chemicals inorganic             \\
			                & alcohol nicotine         & alcohol nicotinyl               \\
			BC5CDR Chemical & dph                      & ddph                            \\
			                & naoh                     & natrolite                       \\
			                & myo inositol 1 phosphate & myo inositol 1 3 6 triphosphate \\
			\noalign{\smallskip}\hline
		\end{tabular}
\end{table}

Table \ref{error_bio} lists the errors in three bioinformatics NEN datasets.
Our proposed model achieves approximately 90\% accuracy for all three datasets.
However, finding the synonyms for short abbreviations such as ``cdm", ``htn", and ``dph" seems relatively harder.
In addition, if there exist longer overlapping strings, the performance of the model is degraded.

\begin{table}
	\caption{Error Analysis on Financial NEN dataset}
	\label{error_finance}
	\centering
	\begin{tabular}{ll}
		\hline\noalign{\smallskip}
		\multicolumn{2}{l}{False Positive}                                        \\
		Entity 1                                     & Entity 2                   \\
		\noalign{\smallskip}\hline\noalign{\smallskip}
		HP Co.                                       & IBM Corporation            \\
		Basel I-derived                              & Basel NSFR                 \\
		Microsoft Corporation ( Microsoft )          & Microsoft Azure            \\
		AT \& T Acquisition                          & AT \& T Corp. ' s ( ATTC ) \\
		Service Cloud                                & Cognitive Cloud Networking \\
		\noalign{\smallskip}\hline\noalign{\smallskip}
		\multicolumn{2}{l}{False Negative}                                        \\
		Entity 1                                     & Entity 2                   \\
		\noalign{\smallskip}\hline\noalign{\smallskip}
		Coca-Cola                                    & Coca-Cola ( r )            \\
		Dodd-Frank ( Orderly Liquidation Authority ) & Dodd-Frank Act             \\
		Gramm-Leach-Bliley Act ( GLB Act )           & GLBA                       \\
		Paris Climate Accords                        & Paris Agreement            \\
		Pfizer Inc. ( Pfizer )                       & Pfizer                     \\
		\noalign{\smallskip}\hline
	\end{tabular}
\end{table}

Financial NEN datasets are constructed using entity pairs.
Our model predicts whether two entity pairs are matched or not.
Table \ref{error_finance} is divided into false positive lists and false-negative lists.
By examine the false-positive lists, entities with similar meanings or with matching strings are often predicted positive while the actual label is negative.

We also examine the false negatives.
Matching named entities with parenthesis and abbreviations is the part where our model's prediction is relatively unstable.
Entity pairs such as ``Paris Climate Accords" and ``Paris Agreement" can be more difficult to predict as positive because the intrinsic meaning of ``Paris Agreement" requires common sense.
Even our model is based on BERT, which captures the semantic meaning from the sentences where named entities are excerpted, using the common sense beyond the information presented in surrounding sentences can be limited.

\subsubsection{Entity Recommendation Result According to Training Progresses}
\label{rec_progression}

\begin{table}
	\caption{Entity Recommendations for Epoch 0, Epoch 1, and Epoch with Highest Accuracy}
	\label{entity_rec}
	\centering
	\resizebox{\textwidth}{!}{\begin{tabular}{llll}
			\hline\noalign{\smallskip}
			      & \multicolumn{3}{l}{NCBI Disease: \textit{\textbf{c2 deficiency}}}                                                                                                                                 \\
			      & Epoch 0                                                                         & Epoch 1                                                & Epoch with Highest Accuracy                            \\
			\noalign{\smallskip}\hline\noalign{\smallskip}
			Top 1 & \underline{\textbf{c2 deficiency}}                                              & \underline{\textbf{c2 deficiency}}                     & \underline{\textbf{c2 deficiency}}                     \\
			Top 2 & c3 deficiency                                                                   & c6 deficiency                                          & \underline{\textbf{c2 deficient}}                      \\
			Top 3 & t2 deficiency                                                                   & c3 deficiency                                          & \underline{\textbf{hereditary c2 deficiency}}          \\
			Top 4 & c5 deficiency                                                                   & \underline{\textbf{c2 deficient}}                      & \underline{\textbf{type ii c2 deficiency}}             \\
			Top 5 & cpox deficiency                                                                 & c4 deficiency                                          & \underline{\textbf{type i c2 deficiency}}              \\
			\noalign{\smallskip}\hline\noalign{\smallskip}
			      & \multicolumn{3}{l}{BC5CDR Disease: \textit{\textbf{failing left ventricle}}}                                                                                                                      \\
			      & Epoch 0                                                                         & Epoch 1                                                & Epoch with Highest Accuracy                            \\
			\noalign{\smallskip}\hline\noalign{\smallskip}
			Top 1 & tumor cerebral ventricle                                                        & dysfunction left ventricular                           & \underline{\textbf{left sided heart failure}}          \\
			Top 2 & cerebral ventricle tumor                                                        & \underline{\textbf{left sided heart failure}}          & \underline{\textbf{heart failure}}                     \\
			Top 3 & tumors cerebral ventricle                                                       & remodeling left ventricular                            & \underline{\textbf{cardiac failure}}                   \\
			Top 4 & syndrome slit ventricle                                                         & hypertrophy left ventricular                           & \underline{\textbf{heart failure left sided}}          \\
			Top 5 & ventricle tumor cerebral                                                        & outflow obstruction left ventricular                   & \underline{\textbf{right sided heart failure}}         \\
			\noalign{\smallskip}\hline\noalign{\smallskip}
			      & \multicolumn{3}{l}{BC5CDR Chemical: \textit{\textbf{vincristine sulfate}}}                                                                                                                        \\
			      & Epoch 0                                                                         & Epoch 1                                                & Epoch with Highest Accuracy                            \\
			\noalign{\smallskip}\hline\noalign{\smallskip}
			Top 1 & \underline{\textbf{vincristine sulfate}}                                        & \underline{\textbf{vincristine sulfate}}               & \underline{\textbf{vincristine sulfate}}               \\
			Top 2 & \underline{\textbf{sulfate vincristine}}                                        & \underline{\textbf{vincristine}}                       & \underline{\textbf{vincristine}}                       \\
			Top 3 & vinblastine sulfate                                                             & voacristine                                            & \underline{\textbf{sulfate vincristine}}               \\
			Top 4 & sulfate vinblastine                                                             & \underline{\textbf{leurocristine}}                     & \underline{\textbf{vincristin}}                        \\
			Top 5 & riboflavin 3 sulfate                                                            & ergocristine                                           & \underline{\textbf{vincristin medac}}                  \\
			\noalign{\smallskip}\hline\noalign{\smallskip}
			      & \multicolumn{3}{l}{Financial NEN: \textit{\textbf{Polo Ralph Lauren Children}}}                                                                                                                   \\
			      & Epoch 0                                                                         & Epoch 1                                                & Epoch with Highest Accuracy                            \\
			\noalign{\smallskip}\hline\noalign{\smallskip}
			Top 1 & Pinky Swear Foundation                                                          & \underline{\textbf{Polo Golf Ralph Lauren}}            & \underline{\textbf{Polo Ralph Lauren}}                 \\
			Top 2 & Bath \& Body Works Canada                                                       & \underline{\textbf{Polo Ralph Lauren}}                 & \underline{\textbf{Polo Ralph Lauren Children, Chaps}} \\
			Top 3 & Ticketmaster North America                                                      & Siemens Medical Solutions USA                          & \underline{\textbf{Polo Golf Ralph Lauren}}            \\
			Top 4 & LIP-BU TAN                                                                      & Mojo Networks, Inc.                                    & Lilly International                                    \\
			Top 5 & Coca-Cola Life                                                                  & \underline{\textbf{Polo Ralph Lauren Children, Chaps}} & \underline{\textbf{Polo / Lauren Company, LP}}         \\
			\noalign{\smallskip}\hline
		\end{tabular}}
\end{table}
As the training epochs increase, recommendations become more accurate.
We randomly selected entities from four datasets we tested.
Top 5 recommendations for the selected entities are provided for epoch 0, epoch 1, and epoch with best result in Section \ref{quantitative_analysis_bio} and Section \ref{quantitative_analysis_finance}.

Table \ref{entity_rec} shows how recommendations change as training progress.
Entities after each dataset are the examples excerpted (c2 deficiency, failing left ventricle, vincristine sulfate, and Polo Ralph Lauren Children), and bold-underlined entities are the entities with the same concept ID as the query entity.
Throughout the datasets, at epoch 0, the recommended entities differ greatly from the concept ID of the query entity.
As the model is trained, the recommendation becomes more accurate in epoch 1.
At the epochs in which the highest accuracy for the datasets is achieved, true synonyms for query entities are successfully selected.

Based on our experiments, our proposed model has the highest precision, recall, F1 score, and accuracy.
Qualitative analysis shows that our proposed model also gives the most robust results.
Our proposed model is most suitable for tasks such as financial named entity normalization automation and preprocessing for various financial NLP tasks.

\section{Conclusion}
\label{conclusion}
We introduce Edge Weight Updating Neural Network for NEN.
NEN to match extracted named entities with homogeneous identity is pivotal for many text mining tasks.
We tested our model on three widely used NEN datasets, NCBI Disease, BC5CDR Disease, and BC5CDR Chemical.
We also generated the NEN dataset for the finance domain.
Next, we verify our model's performance for general NEN applications.

The main contribution of this study are as follows.
Our proposed model successfully links named entities with the same meanings with different surface forms.
The proposed model performs best among previous NEN models.
We test our model not only for bioinformatics datasets in which NEN researches are more active but also for financial NEN datasets.
According to the performance of the NEN corpus in two distinct fields, our proposed model proves the efficacy for general NEN applications.

Similar to many other NEN models, the performance of linking named entities with abbreviations is comparatively lower.
Matching abbreviations more accurately is one of the future works.
The neural network model with our proposed Edge Weight Updating objective function performs better than other models.
Providing the more general guideline for the number of training epochs and increasing the training stability is one of the future research topics.

\section*{acknowledgements}
	This work was supported by National Research Foundation of Korea\\ (2018R1D1A1A02045842).

\section*{Conflict of interest}
This work was supported by National Research Foundation of Korea\\ (2018R1D1A1A02045842).
All authors certify that they have no affiliations with or involvement in any organization or entity with any financial interest or non-financial interest in the subject matter or materials discussed in this manuscript.

\bibliographystyle{unsrtnat}
\bibliography{reference}  %%% Uncomment this line and comment out the ``thebibliography'' section below to use the external .bib file (using bibtex) .

\begin{thebibliography}{50}
\providecommand{\natexlab}[1]{#1}
\providecommand{\url}[1]{\texttt{#1}}
\expandafter\ifx\csname urlstyle\endcsname\relax
  \providecommand{\doi}[1]{doi: #1}\else
  \providecommand{\doi}{doi: \begingroup \urlstyle{rm}\Url}\fi

\bibitem[Devlin et~al.(2018)Devlin, Chang, Lee, and Toutanova]{devlin2018bert}
Jacob Devlin, Ming-Wei Chang, Kenton Lee, and Kristina Toutanova.
\newblock Bert: Pre-training of deep bidirectional transformers for language
  understanding.
\newblock \emph{arXiv preprint arXiv:1810.04805}, 2018.

\bibitem[Cho et~al.(2017)Cho, Choi, and Lee]{cho2017method}
Hyejin Cho, Wonjun Choi, and Hyunju Lee.
\newblock A method for named entity normalization in biomedical articles:
  application to diseases and plants.
\newblock \emph{BMC bioinformatics}, 18\penalty0 (1):\penalty0 451, 2017.

\bibitem[Hanisch et~al.(2005)Hanisch, Fundel, Mevissen, Zimmer, and
  Fluck]{hanisch2005prominer}
Daniel Hanisch, Katrin Fundel, Heinz-Theodor Mevissen, Ralf Zimmer, and Juliane
  Fluck.
\newblock Prominer: rule-based protein and gene entity recognition.
\newblock \emph{BMC bioinformatics}, 6\penalty0 (1):\penalty0 1--9, 2005.

\bibitem[Aronson(2001)]{aronson2001effective}
Alan~R Aronson.
\newblock Effective mapping of biomedical text to the umls metathesaurus: the
  metamap program.
\newblock In \emph{Proceedings of the AMIA Symposium}, page~17. American
  Medical Informatics Association, 2001.

\bibitem[Leaman et~al.(2013)Leaman, Islamaj~Do{\u{g}}an, and
  Lu]{leaman2013dnorm}
Robert Leaman, Rezarta Islamaj~Do{\u{g}}an, and Zhiyong Lu.
\newblock Dnorm: disease name normalization with pairwise learning to rank.
\newblock \emph{Bioinformatics}, 29\penalty0 (22):\penalty0 2909--2917, 2013.

\bibitem[Leaman and Lu(2016)]{leaman2016taggerone}
Robert Leaman and Zhiyong Lu.
\newblock Taggerone: joint named entity recognition and normalization with
  semi-markov models.
\newblock \emph{Bioinformatics}, 32\penalty0 (18):\penalty0 2839--2846, 2016.

\bibitem[Wei and Kao(2011)]{wei2011cross}
Chih-Hsuan Wei and Hung-Yu Kao.
\newblock Cross-species gene normalization by species inference.
\newblock \emph{BMC bioinformatics}, 12\penalty0 (S8):\penalty0 S5, 2011.

\bibitem[Hakenberg et~al.(2011)Hakenberg, Gerner, Haeussler, Solt, Plake,
  Schroeder, Gonzalez, Nenadic, and Bergman]{hakenberg2011gnat}
J{\"o}rg Hakenberg, Martin Gerner, Maximilian Haeussler, Ill{\'e}s Solt, Conrad
  Plake, Michael Schroeder, Graciela Gonzalez, Goran Nenadic, and Casey~M
  Bergman.
\newblock The gnat library for local and remote gene mention normalization.
\newblock \emph{Bioinformatics}, 27\penalty0 (19):\penalty0 2769--2771, 2011.

\bibitem[Rockt{\"a}schel et~al.(2012)Rockt{\"a}schel, Weidlich, and
  Leser]{rocktaschel2012chemspot}
Tim Rockt{\"a}schel, Michael Weidlich, and Ulf Leser.
\newblock Chemspot: a hybrid system for chemical named entity recognition.
\newblock \emph{Bioinformatics}, 28\penalty0 (12):\penalty0 1633--1640, 2012.

\bibitem[Weston et~al.(2019)Weston, Tshitoyan, Dagdelen, Kononova, Trewartha,
  Persson, Ceder, and Jain]{weston2019named}
Leigh Weston, Vahe Tshitoyan, John Dagdelen, Olga Kononova, Amalie Trewartha,
  Kristin~A Persson, Gerbrand Ceder, and Anubhav Jain.
\newblock Named entity recognition and normalization applied to large-scale
  information extraction from the materials science literature.
\newblock \emph{Journal of chemical information and modeling}, 59\penalty0
  (9):\penalty0 3692--3702, 2019.

\bibitem[Suominen et~al.(2013)Suominen, Salanter{\"a}, Velupillai, Chapman,
  Savova, Elhadad, Pradhan, South, Mowery, Jones, et~al.]{suominen2013overview}
Hanna Suominen, Sanna Salanter{\"a}, Sumithra Velupillai, Wendy~W Chapman,
  Guergana Savova, Noemie Elhadad, Sameer Pradhan, Brett~R South, Danielle~L
  Mowery, Gareth~JF Jones, et~al.
\newblock Overview of the share/clef ehealth evaluation lab 2013.
\newblock In \emph{International Conference of the Cross-Language Evaluation
  Forum for European Languages}, pages 212--231. Springer, 2013.

\bibitem[Do{\u{g}}an et~al.(2014)Do{\u{g}}an, Leaman, and Lu]{dougan2014ncbi}
Rezarta~Islamaj Do{\u{g}}an, Robert Leaman, and Zhiyong Lu.
\newblock Ncbi disease corpus: a resource for disease name recognition and
  concept normalization.
\newblock \emph{Journal of biomedical informatics}, 47:\penalty0 1--10, 2014.

\bibitem[Demner-Fushman et~al.(2018)Demner-Fushman, Shooshan, Rodriguez,
  Aronson, Lang, Rogers, Roberts, and Tonning]{demner2018dataset}
Dina Demner-Fushman, Sonya~E Shooshan, Laritza Rodriguez, Alan~R Aronson,
  Francois Lang, Willie Rogers, Kirk Roberts, and Joseph Tonning.
\newblock A dataset of 200 structured product labels annotated for adverse drug
  reactions.
\newblock \emph{Scientific data}, 5:\penalty0 180001, 2018.

\bibitem[Smith et~al.(2008)Smith, Tanabe, nee Ando, Kuo, Chung, Hsu, Lin,
  Klinger, Friedrich, Ganchev, et~al.]{smith2008overview}
Larry Smith, Lorraine~K Tanabe, Rie~Johnson nee Ando, Cheng-Ju Kuo, I-Fang
  Chung, Chun-Nan Hsu, Yu-Shi Lin, Roman Klinger, Christoph~M Friedrich, Kuzman
  Ganchev, et~al.
\newblock Overview of biocreative ii gene mention recognition.
\newblock \emph{Genome biology}, 9\penalty0 (S2):\penalty0 S2, 2008.

\bibitem[Kim et~al.(2009)Kim, Ohta, Pyysalo, Kano, and Tsujii]{kim2009overview}
Jin-Dong Kim, Tomoko Ohta, Sampo Pyysalo, Yoshinobu Kano, and Jun’ichi
  Tsujii.
\newblock Overview of bionlp’09 shared task on event extraction.
\newblock In \emph{Proceedings of the BioNLP 2009 workshop companion volume for
  shared task}, pages 1--9, 2009.

\bibitem[Bossy et~al.(2019)Bossy, Del{\'e}ger, Chaix, Ba, and
  N{\'e}dellec]{bossy2019bacteria}
Robert Bossy, Louise Del{\'e}ger, Estelle Chaix, Mouhamadou Ba, and Claire
  N{\'e}dellec.
\newblock Bacteria biotope at bionlp open shared tasks 2019.
\newblock In \emph{Proceedings of The 5th Workshop on BioNLP Open Shared
  Tasks}, pages 121--131, 2019.

\bibitem[Kol{\'a}rik et~al.(2008)Kol{\'a}rik, Klinger, Friedrich,
  Hofmann-Apitius, and Fluck]{kolarik2008chemical}
Corinna Kol{\'a}rik, Roman Klinger, Christoph~M Friedrich, Martin
  Hofmann-Apitius, and Juliane Fluck.
\newblock Chemical names: terminological resources and corpora annotation.
\newblock In \emph{Workshop on Building and evaluating resources for biomedical
  text mining (6th edition of the Language Resources and Evaluation
  Conference)}, 2008.

\bibitem[Klinger et~al.(2008)Klinger, Kol{\'a}{\v{r}}ik, Fluck,
  Hofmann-Apitius, and Friedrich]{klinger2008detection}
Roman Klinger, Corinna Kol{\'a}{\v{r}}ik, Juliane Fluck, Martin
  Hofmann-Apitius, and Christoph~M Friedrich.
\newblock Detection of iupac and iupac-like chemical names.
\newblock \emph{Bioinformatics}, 24\penalty0 (13):\penalty0 i268--i276, 2008.

\bibitem[Arratia et~al.(2019)Arratia, Belanche, and
  F{\'a}bregues]{arratia2019evaluation}
Argimiro Arratia, Llu{\'\i}s~A Belanche, and Luis F{\'a}bregues.
\newblock An evaluation of equity premium prediction using multiple kernel
  learning with financial features.
\newblock \emph{Neural Processing Letters}, pages 1--18, 2019.

\bibitem[Corba et~al.(2020)Corba, Egrioglu, and Dalar]{corba2020ar}
Burcin~Seyda Corba, Erol Egrioglu, and Ali~Zafer Dalar.
\newblock Ar--arch type artificial neural network for forecasting.
\newblock \emph{Neural Processing Letters}, 51\penalty0 (1):\penalty0 819--836,
  2020.

\bibitem[Gupta et~al.(2020)Gupta, Dengre, Kheruwala, and
  Shah]{gupta2020comprehensive}
Aaryan Gupta, Vinya Dengre, Hamza~Abubakar Kheruwala, and Manan Shah.
\newblock Comprehensive review of text-mining applications in finance.
\newblock \emph{Financial Innovation}, 6\penalty0 (1):\penalty0 1--25, 2020.

\bibitem[Jijkoun et~al.(2008)Jijkoun, Khalid, Marx, and
  De~Rijke]{jijkoun2008named}
Valentin Jijkoun, Mahboob~Alam Khalid, Maarten Marx, and Maarten De~Rijke.
\newblock Named entity normalization in user generated content.
\newblock In \emph{Proceedings of the second workshop on Analytics for noisy
  unstructured text data}, pages 23--30, 2008.

\bibitem[Sun et~al.(2012)Sun, Lin, Liu, Liu, and Sha]{sun2012product}
Chengjie Sun, Lei Lin, Ming Liu, Bingquan Liu, and Xuejun Sha.
\newblock A product named entity normalization method based on entity
  relations.
\newblock In \emph{2012 8th International Conference on Information Science and
  Digital Content Technology (ICIDT2012)}, volume~1, pages 166--169. IEEE,
  2012.

\bibitem[Francis et~al.(2019)Francis, Van~Landeghem, and
  Moens]{francis2019transfer}
Sumam Francis, Jordy Van~Landeghem, and Marie-Francine Moens.
\newblock Transfer learning for named entity recognition in financial and
  biomedical documents.
\newblock \emph{Information}, 10\penalty0 (8):\penalty0 248, 2019.

\bibitem[Mueller and Thyagarajan(2016)]{mueller2016siamese}
Jonas Mueller and Aditya Thyagarajan.
\newblock Siamese recurrent architectures for learning sentence similarity.
\newblock In \emph{Proceedings of the AAAI Conference on Artificial
  Intelligence}, volume~30, 2016.

\bibitem[Ranasinghe et~al.(2019)Ranasinghe, Orasan, and
  Mitkov]{ranasinghe2019semantic}
Tharindu Ranasinghe, Constantin Orasan, and Ruslan Mitkov.
\newblock Semantic textual similarity with siamese neural networks.
\newblock In \emph{Proceedings of the International Conference on Recent
  Advances in Natural Language Processing (RANLP 2019)}, pages 1004--1011,
  2019.

\bibitem[Liu et~al.(2018)Liu, Zhang, Niu, Lin, Lai, and Xu]{liu2018matching}
Bang Liu, Ting Zhang, Di~Niu, Jinghong Lin, Kunfeng Lai, and Yu~Xu.
\newblock Matching long text documents via graph convolutional networks.
\newblock \emph{arXiv preprint arXiv:1802.07459}, pages 2793--2799, 2018.

\bibitem[Krivosheev et~al.(2020)Krivosheev, Atzeni, Mirylenka, Scotton, and
  Casati]{krivosheev2020siamese}
Evgeny Krivosheev, Mattia Atzeni, Katsiaryna Mirylenka, Paolo Scotton, and
  Fabio Casati.
\newblock Siamese graph neural networks for data integration.
\newblock \emph{arXiv preprint arXiv:2001.06543}, 2020.

\bibitem[Neculoiu et~al.(2016)Neculoiu, Versteegh, and
  Rotaru]{neculoiu2016learning}
Paul Neculoiu, Maarten Versteegh, and Mihai Rotaru.
\newblock Learning text similarity with siamese recurrent networks.
\newblock In \emph{Proceedings of the 1st Workshop on Representation Learning
  for NLP}, pages 148--157, 2016.

\bibitem[Niu et~al.(2019)Niu, Yang, Zhang, Sun, and Zhang]{niu2019multi}
Jinghao Niu, Yehui Yang, Siheng Zhang, Zhengya Sun, and Wensheng Zhang.
\newblock Multi-task character-level attentional networks for medical concept
  normalization.
\newblock \emph{Neural Processing Letters}, 49\penalty0 (3):\penalty0
  1239--1256, 2019.

\bibitem[Mulang' et~al.(2020)Mulang', Singh, Prabhu, Nadgeri, Hoffart, and
  Lehmann]{mulang2020evaluating}
Isaiah~Onando Mulang', Kuldeep Singh, Chaitali Prabhu, Abhishek Nadgeri,
  Johannes Hoffart, and Jens Lehmann.
\newblock Evaluating the impact of knowledge graph context on entity
  disambiguation models.
\newblock In \emph{Proceedings of the 29th ACM International Conference on
  Information \& Knowledge Management}, pages 2157--2160, 2020.

\bibitem[D’Souza and Ng(2015)]{d2015sieve}
Jennifer D’Souza and Vincent Ng.
\newblock Sieve-based entity linking for the biomedical domain.
\newblock In \emph{Proceedings of the 53rd Annual Meeting of the Association
  for Computational Linguistics and the 7th International Joint Conference on
  Natural Language Processing (Volume 2: Short Papers)}, pages 297--302, 2015.

\bibitem[Li et~al.(2017)Li, Chen, Tang, Wang, Xu, Wang, and Huang]{li2017cnn}
Haodi Li, Qingcai Chen, Buzhou Tang, Xiaolong Wang, Hua Xu, Baohua Wang, and
  Dong Huang.
\newblock Cnn-based ranking for biomedical entity normalization.
\newblock \emph{BMC bioinformatics}, 18\penalty0 (11):\penalty0 79--86, 2017.

\bibitem[Wright(2019)]{wright2019normco}
Dustin Wright.
\newblock \emph{NormCo: Deep disease normalization for biomedical knowledge
  base construction}.
\newblock PhD thesis, UC San Diego, 2019.

\bibitem[Phan et~al.(2019)Phan, Sun, and Tay]{phan2019robust}
Minh~C Phan, Aixin Sun, and Yi~Tay.
\newblock Robust representation learning of biomedical names.
\newblock In \emph{Proceedings of the 57th Annual Meeting of the Association
  for Computational Linguistics}, pages 3275--3285, 2019.

\bibitem[Ji et~al.(2020)Ji, Wei, and Xu]{ji2020bert}
Zongcheng Ji, Qiang Wei, and Hua Xu.
\newblock Bert-based ranking for biomedical entity normalization.
\newblock \emph{AMIA Summits on Translational Science Proceedings},
  2020:\penalty0 269, 2020.

\bibitem[Sung et~al.(2020)Sung, Jeon, Lee, and Kang]{sung2020biomedical}
Mujeen Sung, Hwisang Jeon, Jinhyuk Lee, and Jaewoo Kang.
\newblock Biomedical entity representations with synonym marginalization.
\newblock \emph{arXiv preprint arXiv:2005.00239}, 2020.

\bibitem[Kim et~al.(2019)Kim, Kim, Kim, and Yoo]{kim2019edge}
Jongmin Kim, Taesup Kim, Sungwoong Kim, and Chang~D Yoo.
\newblock Edge-labeling graph neural network for few-shot learning.
\newblock In \emph{Proceedings of the IEEE/CVF Conference on Computer Vision
  and Pattern Recognition}, pages 11--20, 2019.

\bibitem[Kullback and Leibler(1951)]{kullback1951information}
Solomon Kullback and Richard~A Leibler.
\newblock On information and sufficiency.
\newblock \emph{The annals of mathematical statistics}, 22\penalty0
  (1):\penalty0 79--86, 1951.

\bibitem[Lee et~al.(2020)Lee, Yoon, Kim, Kim, Kim, So, and
  Kang]{lee2020biobert}
Jinhyuk Lee, Wonjin Yoon, Sungdong Kim, Donghyeon Kim, Sunkyu Kim, Chan~Ho So,
  and Jaewoo Kang.
\newblock Biobert: a pre-trained biomedical language representation model for
  biomedical text mining.
\newblock \emph{Bioinformatics}, 36\penalty0 (4):\penalty0 1234--1240, 2020.

\bibitem[Loshchilov and Hutter(2017)]{loshchilov2017decoupled}
Ilya Loshchilov and Frank Hutter.
\newblock Decoupled weight decay regularization.
\newblock \emph{arXiv preprint arXiv:1711.05101}, 2017.

\bibitem[Li et~al.(2016)Li, Sun, Johnson, Sciaky, Wei, Leaman, Davis,
  Mattingly, Wiegers, and Lu]{li2016biocreative}
Jiao Li, Yueping Sun, Robin~J Johnson, Daniela Sciaky, Chih-Hsuan Wei, Robert
  Leaman, Allan~Peter Davis, Carolyn~J Mattingly, Thomas~C Wiegers, and Zhiyong
  Lu.
\newblock Biocreative v cdr task corpus: a resource for chemical disease
  relation extraction.
\newblock \emph{Database}, 2016, 2016.

\bibitem[Davis et~al.(2009)Davis, Murphy, Saraceni-Richards, Rosenstein,
  Wiegers, and Mattingly]{davis2009comparative}
Allan~Peter Davis, Cynthia~G Murphy, Cynthia~A Saraceni-Richards, Michael~C
  Rosenstein, Thomas~C Wiegers, and Carolyn~J Mattingly.
\newblock Comparative toxicogenomics database: a knowledgebase and discovery
  tool for chemical--gene--disease networks.
\newblock \emph{Nucleic acids research}, 37\penalty0 (suppl\_1):\penalty0
  D786--D792, 2009.

\bibitem[Sang and De~Meulder(2003)]{sang2003introduction}
Erik~F Sang and Fien De~Meulder.
\newblock Introduction to the conll-2003 shared task: Language-independent
  named entity recognition.
\newblock \emph{arXiv preprint cs/0306050}, 2003.

\bibitem[Si et~al.(2019)Si, Wang, Xu, and Roberts]{si2019enhancing}
Yuqi Si, Jingqi Wang, Hua Xu, and Kirk Roberts.
\newblock Enhancing clinical concept extraction with contextual embeddings.
\newblock \emph{Journal of the American Medical Informatics Association},
  26\penalty0 (11):\penalty0 1297--1304, 2019.

\bibitem[Mondal et~al.(2020)Mondal, Purkayastha, Sarkar, Goyal, Pillai,
  Bhattacharyya, and Gattu]{mondal2020medical}
Ishani Mondal, Sukannya Purkayastha, Sudeshna Sarkar, Pawan Goyal, Jitesh
  Pillai, Amitava Bhattacharyya, and Mahanandeeshwar Gattu.
\newblock Medical entity linking using triplet network.
\newblock \emph{arXiv preprint arXiv:2012.11164}, 2020.

\bibitem[Hoffer and Ailon(2015)]{hoffer2015deep}
Elad Hoffer and Nir Ailon.
\newblock Deep metric learning using triplet network.
\newblock In \emph{International workshop on similarity-based pattern
  recognition}, pages 84--92. Springer, 2015.

\bibitem[Levenshtein(1966)]{levenshtein1966binary}
Vladimir~I Levenshtein.
\newblock Binary codes capable of correcting deletions, insertions, and
  reversals.
\newblock In \emph{Soviet physics doklady}, volume~10, pages 707--710, 1966.

\bibitem[Kipf and Welling(2016)]{kipf2016semi}
Thomas~N Kipf and Max Welling.
\newblock Semi-supervised classification with graph convolutional networks.
\newblock \emph{arXiv preprint arXiv:1609.02907}, 2016.

\bibitem[Schuster and Paliwal(1997)]{schuster1997bidirectional}
Mike Schuster and Kuldip~K Paliwal.
\newblock Bidirectional recurrent neural networks.
\newblock \emph{IEEE transactions on Signal Processing}, 45\penalty0
  (11):\penalty0 2673--2681, 1997.

\end{thebibliography}





% \end{thebibliography}

\end{document}